\def\year{2021}\relax
\documentclass[letterpaper]{article} 
\usepackage{aaai21}  
\usepackage{times}  
\usepackage{helvet} 
\usepackage{courier}  
\usepackage[hyphens]{url}  
\usepackage{graphicx} 
\usepackage[dvipsnames, svgnames, x11names]{xcolor}
\usepackage{natbib}  
\usepackage{caption} 
\title{How to Train Your Agent to Read and Write}

\usepackage[utf8]{inputenc} 
\usepackage[T1]{fontenc}    
\usepackage{url}            
\usepackage{booktabs}       
\usepackage{amsfonts}       
\usepackage{nicefrac}      
\usepackage{microtype}      
\usepackage{graphicx}
\usepackage{amsmath}
\usepackage{wrapfig}
\usepackage{bm}
\usepackage{xcolor}
\usepackage{floatrow}
\usepackage{caption}
\usepackage{enumitem}
\usepackage{mathrsfs}
\usepackage{multirow}
\usepackage{lscape}
\usepackage[switch]{lineno}
\newfloatcommand{capbtabbox}{table}[][\FBwidth]

\def\xgh{\textcolor{black}}
\def\hmg{\textcolor{black}}

\def\wzq{\textcolor{black}}

\def\dcr{\textcolor{black}}

\def\eg{\emph{e.g., }} 

\def\ie{\emph{i.e., }}

\def\wrt{\emph{w.r.t. }}

\begin{document}
\author{
\\
Li Liu,$^{1,2}$\thanks{Authors contributed equally.} Mengge He,$^{1*}$ Guanghui Xu,$^{1}$ Mingkui Tan,$^{1,4}$\thanks{Corresponding author.} Qi Wu$^{3}$
\\
}
\affiliations{
$^{1}$ School of Software Engineering, South China University of Technology, \\
$^{2}$ Pazhou Laboratory, 
$^{3}$ University of Adelaide, \\
$^{4}$ Key Laboratory of Big Data and Intelligent Robot, Ministry of Education  \\
\{seliushiya, semenggehe, sexuguanghui\}@mail.scut.edu.cn, mingkuitan@scut.edu.cn, qi.wu01@adelaide.edu.au
}
\maketitle
\begin{abstract}
Reading and writing research papers is one of the most privileged abilities that a qualified researcher should master. However, it is difficult for new researchers (\eg{students}) to fully {grasp} this ability.
It would be fascinating if we could train an intelligent agent to help people read and summarize papers, and perhaps even discover and exploit the potential knowledge clues to write novel papers. 
Although there have been existing works focusing on summarizing (\emph{i.e.}, reading) the knowledge in a given text or generating (\emph{i.e.}, writing) a text based on the given knowledge, the ability of simultaneously reading and writing is still under development. Typically, this requires an agent to fully understand the knowledge from the given text materials and generate correct and fluent novel paragraphs, which is very challenging in practice.
In this paper, we propose a Deep ReAder-Writer (DRAW) network, which consists of a \textit{Reader} that can extract knowledge graphs (KGs) from input paragraphs and discover potential knowledge,
a graph-to-text \textit{Writer} that generates a novel paragraph, and a \textit{Reviewer} that reviews the generated paragraph from three different aspects. Extensive experiments show that our DRAW network outperforms considered baselines and several state-of-the-art methods on AGENDA and M-AGENDA datasets. Our code and supplementary are released at https://github.com/menggehe/DRAW.
\end{abstract}

\section{Introduction}

\hmg{Currently,}
hundreds of papers are published online every day even 
\hmg{on small topics.}
However, a study~\cite{Wang2019PaperRobotID} shows that US scientists can only read 264 papers per year on average.
\hmg{Thus, researchers are exhausted by following the sharply increased numbers of papers, much less to understanding the research and coming up with new ideas to write novel papers~\cite{Gopen1990TheSO,Buenz2019EssentialEF}.}
In practice, writing novel papers requires not only the abilities of reading and reasoning but also the ability of creative thinking, which is nontrivial for most fresh researchers~\cite{DBLP:conf/aaai/XiaoWHJ20}. It would be fantastic if an agent could help people, especially \hmg{new} researchers, to read and write. However, building such an agent encounters several challenges.

First, to understand multiple related works, the agent needs to capture complex logic in the related works, which is nontrivial.
Several knowledge extraction methods~\cite{zhang-etal-2006-composite,gerber-chai-2010-beyond,Yoshikawa2010CoreferenceBE} achieve it by identifying entities in the texts, extracting the relationships between these entities, and representing them as a knowledge graph (KG).
However, 
\hmg{they have trouble in discovering potential connections among these entities, which hampers a comprehensive understanding of related works.}

\begin{figure}[t]
    \includegraphics[width=1 \linewidth]{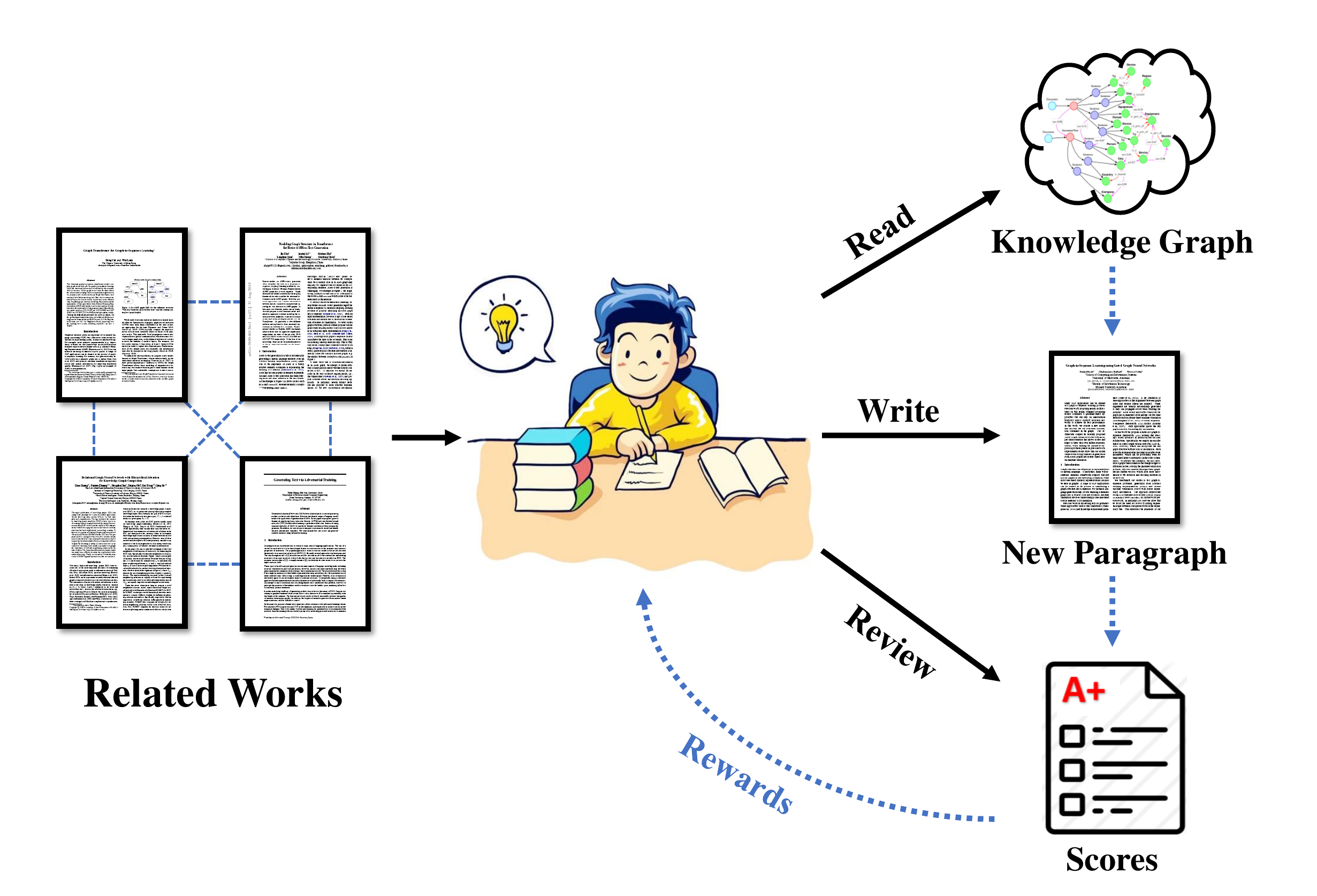}
    \caption{An intuitive understanding of our DRAW network. First, the DRAW network reads multiple related works and discovers potential knowledge among them. And then, it writes a new paragraph based on knowledge graph. Last, it reviews the output and uses feedback rewards to improve the quality of writing.}
    \label{fig:simple_framework}
\end{figure}

Second, after generating a KG, the agent is then required to decode a fluent novel paragraph from the KG. In practice, however, how to evaluate the quality of the generated texts accurately is still an open problem. Existing methods~\cite{KoncelKedziorski2019TextGF,Wang2019PaperRobotID}
adopt the teacher-forcing scheme that aims to match the tokens in the generated texts to the tokens in the target texts.
However, these methods only focus on token-level matching while ignoring sentence-level and graph-level evaluation of the generated texts.

In this paper, we propose a method named 
Deep ReAder-Writer (DRAW). Our DRAW network is able to read multiple texts, discover potential knowledge, and then write a novel paragraph. 
From Figure~\ref{fig:simple_framework}, the DRAW network consists of three modules: ~\ie \textit{Reader}, \textit{Writer} and \textit{Reviewer}.
Specifically, the \textit{Reader} first extracts KGs from the research texts and discovers potential knowledge to enrich the KGs.
The \textit{Reader} considers the multi-hop neighborhood to predict new links among conceptual nodes. 
Then, the \textit{Writer} writes a novel paragraph to describe the main idea of the enriched KGs using a graph attention network, which aggregates the global and local graph information.
Inspired by the review process of research papers, we further propose a \textit{Reviewer} \dcr{module} to evaluate the quality of the generated paragraphs and return rewards as feedback signals to refine the \textit{Writer}. To be specific, given a generated paragraph, the \textit{Reviewer} will output three feedback signals, including (1) \hmg{a} quality reward, which reflects the metric scores of the generated paragraph; (2) \hmg{an} adversarial reward, which denotes the probability of the generated paragraph passing the Turing test; and (3) \hmg{an} alignment reward, which represents the matching score between the generated paragraphs and the enriched KGs.
In this way, the \textit{Writer} {is able} to write better paragraphs {that} clearly represent the key idea of the enriched KGs.

In summary, our main contributions are threefold:

\begin{itemize}
    \item We propose a Deep ReAder-Writer (DRAW) network  that reads multiple
    research texts
    and then discovers potential knowledge to write a novel paragraph covering the key idea of the source inputs.

    \item 
    We propose a feedback mechanism to review whether the generated paragraph is consistent with the enriched KG, and whether the generated 
    paragraph is human written, thereby greatly improving the quality of paragraph generation.
    
    \item Extensive experiments show that our \textit{Writer}-\textit{Reviewer} leads to significant improvements in \hmg{the} KGs-to-text generation task and outperforms the state-of-the-art methods.

\end{itemize}

\section{Related Work} 
\subsubsection{Automatic writing.}

PaperRobot~\cite{Wang2019PaperRobotID} performs as an automatic research assistant to incrementally write 
\hmg{to}
chemical-related research datasets. 
It enriches KGs by predicting links of input papers' KGs. According to a given title, it then selects several entities that are related to the title in enriched KGs to generate texts. 
However, PaperRobot neglects to consider the multi-hop neighborhood to predict links, which is very important for capturing potential relationships.
In addition, the generated texts do not \hmg{closely} align with the KGs.
To address this, we use a graph attention network to consider the multi-hop neighborhood, capturing the complex and hidden information that is inherently implicit in the neighborhood. 
Moreover, we design a \textit{Reviewer} to measure the quality of the generated text from different dimensions to \hmg{effectively} align with the {KGs}.
In particular, our DRAW network is different from the multi-document summary~\cite{Ling2013Multi}, which compresses the lengthy document content into several relatively short paragraphs. We not only extract important knowledge but also discover potential knowledge from multiple paragraphs by predicting links and \hmg{writing} a novel paragraph.

\subsubsection{Link prediction.}

Some translation-based approaches~\cite{NIPS2013_5071,wang_zhen,Lin2015LearningEA} are widely used in link prediction but result in poor {representation ability}.
Recently, CNN based models \cite{Dettmers2018Convolutional2K, Nguyen2018ANE} have been proposed for relation prediction.
{These methods only focus on the entity and its neighborhood while not considering the relationships among these nodes.
Other methods~\cite{Kipf2017SemiSupervisedCW,inbook} take the relationships among the entities and their 1-hop neighbors into consideration. However, they still omit the information from multi-hop neighborhood.}
Instead, we propose a \textit{Reader} module to capture semantic information of the multi-hop neighborhood in the KG.

\subsubsection{Graph-to-Text task.} 
Graph-to-Text is an active research area. Some works generate texts based on structured knowledge~\cite{GTR-LSTM, SQLtoTextGW, nie-etal-2018-operation}, while several neural graph-to-text models use different encoders based on GNN~\cite{inproceedings,Guo,Huang2020Location} and Transformer~\cite{Vaswani2017AttentionIA} architectures to learn graph representations. \citeauthor{KoncelKedziorski2019TextGF} proposes a novel graph transformer encoder, {which} leverages the topological structure of KGs to generate texts. 
However, it ignores the global graph information, which is important for text generation. 
To solve this, \citeauthor{ribeiro2020modeling} introduce a novel architecture that aggregates both global and local graph information to generate texts. 
However, such an encoder-decoder framework presents some problems such as word repetition and lack of diversity. 
To solve these issues, we propose a \textit{Reviewer} module to review the generated paragraphs and refine the quality of paragraphs using feedback rewards.
Our \textit{Reviewer} consists of three modules to review and evaluate whether the generated paragraphs are real and to align with the given KGs, in order to improve the text generation ability.

\section{Proposed Method}

\begin{figure*}[t]
\centering
{
\includegraphics[width=0.95\linewidth, ]{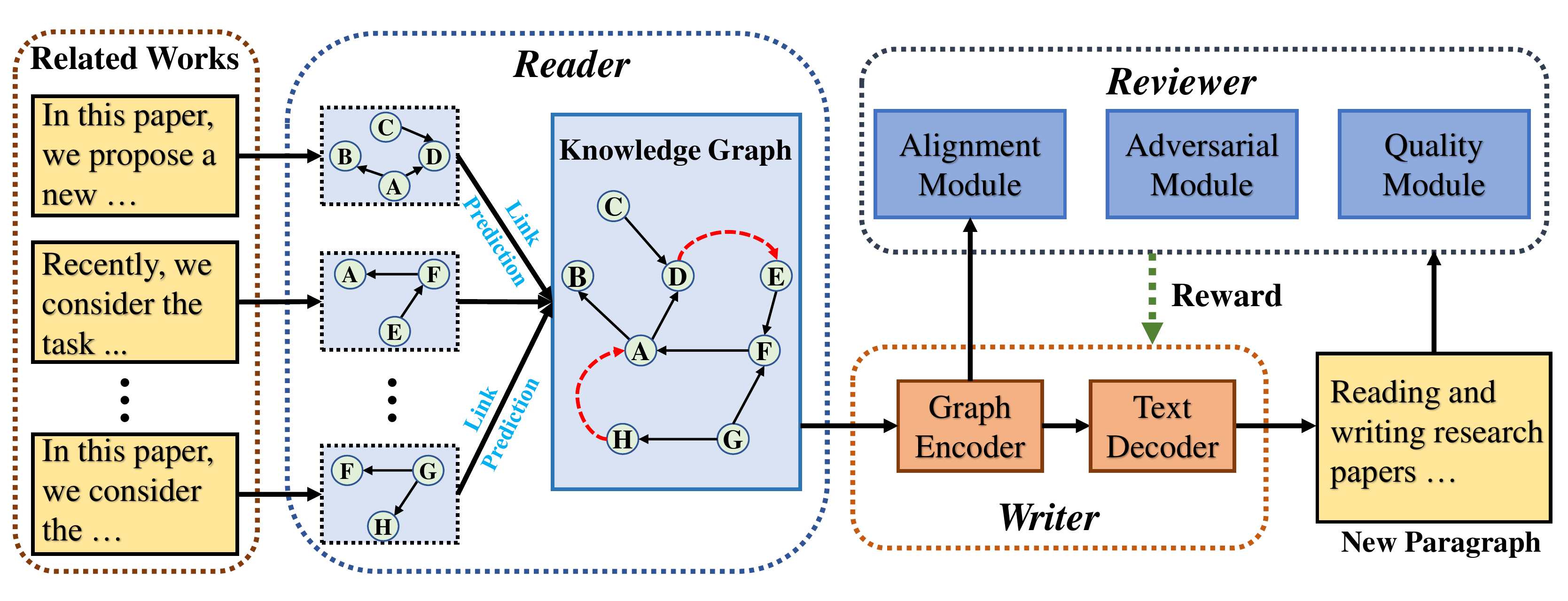}
\caption{An overview of our Deep ReAder-Writer (DRAW) Network. The DRAW network consists of three modules, namely \textit{Reader}, \textit{Writer} and \textit{Reviewer}. Given multiple related works, the \textit{Reader} first extracts knowledge to construct initial knowledge graphs (KGs) and performs link prediction to enrich KGs. Based on the enriched KGs, the \textit{Writer} captures global and local topology information using a graph encoder and generates a novel paragraph with a text decoder. In particular, the \textit{Reviewer} employs three feedback modules to measure the quality of the generated paragraph.}
\label{fig:overview_framework}
}
\end{figure*}

In this paper, we focus on generating novel paragraphs via reading multiple AI-related paragraphs. To this end, we propose a Deep ReAder-Writer (DRAW) network that
consists of three modules, namely \textit{Reader}, \textit{Writer}, and \textit{Reviewer} as shown in Figure~\ref{fig:overview_framework}. 
To understand and sort out the textual logic of given paragraphs, the \textit{Reader} first `reads' and extracts {knowledge graphs} (KGs) from them. And then, considering the multi-hop neighborhood, the \textit{Reader} 
predicts new links between conceptual nodes, namely potential knowledge, to enrich the KGs.
The \textit{Writer} adopts a graph encoder to encode the rich semantic information in KGs, and delivers it to a text encoder to generate a novel paragraph.
{Inspired by the adversarial learning \cite{cao2019multi, wang2018graphgan, cao2020pami, Chen2020Generating}, }
we also devise a \textit{Reviewer} to evaluate the quality of the generated paragraph, which serves as a feedback signal to refine the \textit{Writer}.
We \hmg{relate} the details of these modules in the following sections.

\subsection{Reader: Text-to-Graph}
To extract the textual logic from the given related paragraphs, we use the standard SciIE~\cite{Luan2018MultiTaskIO}, a science domain information extraction system 
{to constrcu knowledge graphs}
Specifically, the output of the SciIE system is a list of triplets, where each triplet consists of two entities and the corresponding relation. 
The knowledge graph denoted as
$\mG_{I}=\{\mV, \mR\}$, 
where $ \mV = \{ \bar{\bv}_i\}^{N}_{i=1} $ is the node set, $ \mR = \{\bar{\br}_{ij}\}^{N}_{i,j=1} $ is the edge set and $N$ represents the number of nodes.
$\mV$ and $\mR$ represent the extracted entities and the relations, respectively.
However, the initial knowledge graph $\mG_{I}$ does not exploit potential knowledge.
To address this, we perform a link prediction to predict new links between entities based on the initial KGs.

\subsubsection{Link prediction.}
Given KGs $\mG_{I}$, we obtain the entity embedding $\bar{\bv}_i \in \mmR^{d}$ and relation embedding $\bar{\br}_{ij} \in \mmR^{d}$ with two separated embedding layers, where $d$ is the feature dimension.
Formally, given entity embedding $\bar{\bv}_i$, $\bar{\bv}_j$ and relation embedding $\bar{\br}_{ij}$ between them, the triplet is represented by $(\bar{\bv}_i, \bar{\br}_{ij}, \bar{\bv}_j)$.
To aggregate more information, we introduce auxiliary edges between one entity and its n-hop neighborhood. For the entity and its n-hop neighborhood, we sum the embeddings of all the relations in the path between them as the auxiliary relation embedding.
We apply a linear transformation to update the entity representation $\widetilde{\bv}_i \in \mmR^{d}$ by:
\begin{equation}
    \begin{aligned}
    \label{eq:update_entity}
    \widetilde{\bv}_i = \bW_{1} [\bar{\bv}_i, \ \bar{\br}_{ij}, \ \bar{\bv}_j],
    \end{aligned}
\end{equation}
where $\bW_1$ is a trainable parameter and $[\cdot,\cdot]$ denotes the concatenation operation. 
A particular entity $\bv_i$ may be involved in multiple triplets and its neighborhood can be denoted as $\{ \widetilde{\bv}_i^{k} \}$, where $\widetilde{\bv}_i^{k}$ denotes the $k$-th neighborhood of \hmg{the} $i$-th entity. 
To learn the importance of each triplet for the entity, we apply a self-attention to calculate attention weights 
as follows:
\begin{equation}
    \begin{aligned}
    \widehat{a}_k =  \frac{\exp \left( \widetilde{\bv}_i^{k} \right)}{\sum_{k}  \exp \left( \widetilde{\bv}_i^{k} \right) },
\end{aligned}
\end{equation}
With the help of the attention weights, we update feature $\bv_i \in \mmR^{d}$ by fusing the information from its neighborhood, \ie
\begin{equation}
    \begin{aligned}
    \bv_i = \bW_{2} \ \widetilde{\bv}_i + \sigma \left(\sum\nolimits_{k} \widehat{a}_k \ \widetilde{\bv}_i^{k}\right),
\end{aligned}
\end{equation}
where $\bW_2$ is a trainable parameter and $\sigma$ is the Sigmoid function.

Based on the original relation feature 
$\bar{\br}_{ij}$, 
we apply a linear transformation to obtain the updated relation embedding $r_{ij} \in \mmR^{d}$.
After updating the node and relation embeddings, we need to 
determine whether there is a relationship between two given entities. An intuitive way is to calculate the probability for each triple. Following ConvKB~\cite{Nguyen2018ANE}, we train a scoring function to perform the relation prediction as follows:
\begin{equation}
    \begin{aligned}
    \label{eq:relation_score}
    s_m = \mathrm{FC}([\bv_i, \br_m, \bv_j] * \mathbf{\Omega}) ,
    \end{aligned}
\end{equation}
where $*$ denotes a convolution operation, {$\mathbf{\Omega} \in \mmR^{1 \times 3}$} is a set of convolution filters, and $\mathrm{FC}(\cdot)$ is a linear transformation layer. Following~\cite{Nathani2019LearningAE}, we assign a score $s_m$ to the triplet $(\bv_i, \br_m, \bv_j)$ in Eqn.(\ref{eq:relation_score}), which indicates the probability that the triplet holds. 
For each entity, we first traverse all entities and relationships to construct triples, and then we select the triplet with the highest score as the new link.
In this way, the \textit{Reader} can capture potential relations between different nodes and derive a new graph $\mG_{P}$.
Finally, we denote the enriched knowledge graph as $\mG= \mG_{I} \cup \mG_{P}$.

\subsection{Writer: Graph-to-Text}
Based on the enriched graph $\mG$ with $N$ entities, we propose a \textit{Writer} to generate novel paragraphs, 
which consists of a graph encoder and a text decoder.
Specifically, the \textit{Writer} first uses the graph encoder to extract the knowledge representations and then writes a new paragraph with the text decoder~\cite{Vaswani2017AttentionIA}.
\subsubsection{Graph encoder.}
A comprehensive understanding of a KG $\mG$ is the first step to generate the desired paragraph. However, it is difficult to directly capture rich semantic information in the knowledge graph $\mG$. 
To address this,
we extract the knowledge representations within two sub-encoders,~\ie~global-graph encoder and local-graph encoder.
Following CGE-LW \cite{ribeiro2020modeling}, we integrate global context information and local topology information to generate new paragraphs.

To aggregate global context information, we first concatenate all of the node features $\bv$ and feed them into the global-graph encoder $\Psi$ as follows:
\begin{equation}
    \begin{aligned}
    \label{global_encoder}
    \left [ \widehat{\bv}_1,\dots, \widehat{\bv}_N \right ] = \Psi ([\bv_1,\dots, \bv_N]),
    \end{aligned}
\end{equation}
where $\Psi$ is a standard Transformer encoder~\cite{Vaswani2017AttentionIA}, which contains multi-head self-attention layers and feed-forward networks. In the global-graph encoder, we treat the knowledge graphs $\mG$ as a fully connected graph without labeled edges. Based on the self-attention mechanism, the global-graph encoder is suitable for discovering the global correlation between nodes. Each node $\widehat{\bv}_i \in \mmR^{d}$ has the ability to capture all nodes' information.

To better represent the interaction between nodes, we need to build local relations between each node and its neighborhood. However, the global-graph encoder does not explicitly consider such graph topology information. To address this, we use the local-graph encoder to model the local relations. 
For each node, we first calculate attention weights for its adjacent nodes since the different 
\hmg{types of relationships have considerable discrepancies }
in impact when fusing information. Based on the attention weights, we obtain the hidden node features $\widehat{\bh}$ by
\begin{equation}
    \begin{aligned}
  \widehat{\bh}_i &= \sum_{j \in \mN_i} a_{ij} \ \br_{ij} \ \widehat{\bv}_j, ~~~\mathrm{where}\\
     \widehat{\br}_{ij} &= \mathrm{ReLU} (\br_{ij} \ \bW_3 \ [\widehat{\bv}_{i}, \widehat{\bv}_{j}]),\\
    a_{ij} &=  \frac{\exp (\widehat{\br}_{ij})}{\sum_{j \in \mN_i}  \exp (\widehat{\br}_{ij}) }.
    \end{aligned}
\end{equation}
Here, $\bW_3$ denotes the model parameters and  $\widehat{\bh}_i$ denotes the hidden features which encode the local interaction between the $i$-th node and its neighborhood.
\hmg{$\mN_i$ denotes the neighbourhood of the $i$-th node.
We also perform the multi-head attention operation to learn structural information from different perspectives.}
\hmg{we employ a GRU~\cite{Cho2014LearningPR} to merge local information between different layers as follows:}

\begin{equation}
    \begin{aligned}
    \bh_i = \mathrm{GRU} (\widehat{\bv}_i, \widehat{\bh}_i),
    \end{aligned}
\end{equation}
where the final node representation $\bh_i \in \mmR^{d}$.

\subsubsection{Text decoder.}

Based on the node representations $\bH=\{\bh_i\}_{i=1}^{N}$, we use the standard Transformer decoder~\cite{Vaswani2017AttentionIA} to generate a novel paragraph $\tau$ with $T$ words in an auto-regression manner. At each step $t$, the text decoder consumes the previously generated tokens as additional input and outputs a probability $\bp_t$ over candidate vocabularies. 
We train the \textit{Writer} with supervised learning as follows:
\begin{equation}
    \begin{aligned}
    \mathcal{L}_{SL} = - \sum_{t=1}^{T} \by_t \ \mathrm{log} (\bp_t),
    \label{Eq:SL loss}
    \end{aligned}
\end{equation}
where $\by_t$ is the ground-truth one-hot vector at step $t$ and \hmg{generates words} by selecting the element with the highest score at this step. In practice, the text decoder also can use other sequence generation models, such as LSTM~\cite{LSTM} 
and so on.

\subsection{Reviewer: Feedback Rewards}

The encoder-decoder framework
has made great progress in many sequence generation tasks, including text summarization and image captioning. Nevertheless, it suffers from some problems. 
For each training sample, such a framework tends to use only one word as ground-truth at each generation step, even if other candidate words are also reasonable. \hmg{This} leads to a lack of diversity in the generated text.
Moreover, the language is so complex that it requires us to evaluate the quality of the generated paragraph from different dimensions, such as grammatical correctness, topic relevance, language and logic coherence, etc. Inspired by the review process of a research paper, we propose a \textit{Reviewer} module to review the generated paragraph from different dimensions. The output of \textit{Reviewer} can be used as an auxiliary training signal to optimize the \textit{Writer}, which is similar to researchers further polishing the paper based on reviews.

Specifically, we design three feedback rewards in the \textit{Reviewer}. 
{First, we use the metric scores of the generated paragraph as a reward to meet the rules of these metrics. }
Second, we train a Turing-Test discriminator to determine whether the paragraph is generated by an agent or written by a human, which draws on the idea of adversarial training and requires the paragraph to conform to the natural language specification. 
Third, we design an alignment module to align the generated paragraphs and the corresponding enriched knowledge graphs, which ensures the correctness and completeness of the generated texts.
Different from teacher-forcing methods, the \textit{Reviewer} focuses on sentence-level and graph-level alignment.
Given a generated paragraph, however, the above evaluation processes are non-differentiable. As discussed in SeqGAN~\cite{Yu2017SeqGANSG}, the discrete signals 
are limited in passing the gradient update from the \textit{Reviewer} to the \textit{Writer}. To address this, we denote the outputs of \textit{Reviewer} as rewards $R$ and maximize expectation rewards $\mmE [\cdot]$ via reinforcement learning. Formally, the goal of \textit{Reviewer} can be represented by
    $\mathrm{max} \ \mmE_{P(\tau;\theta)} [R(\tau)]$,
where $\theta$ denotes the trainable parameters of our model, and $\tau$ is the paragraph generated by the \textit{Writer} based on the generation probability $P$ \wrt~$\theta$. Specifically, the reward function is denoted as
\begin{equation}
    \begin{aligned}
    R(\tau)=R_1 + \lambda_{AR} \ R_2 + \lambda_{MR} \ R_3,
    \label{eq:reward}
    \end{aligned}
\end{equation}
where $R_{1}$, $R_{2}$, \hmg{and} $R_{3}$  correspond to the three modules of \hmg{the} \textit{Reviewer}. $\lambda_{AR}$ and $\lambda_{MR}$ control the contribution of the corresponding reward. Following policy gradient methods~\cite{Williams92simplestatistical,Schulman2017ProximalPO}, we can solve the above problem in batch training as follows:
\begin{equation}
    \begin{aligned}
    \label{eq:pg}
    \mathcal{L}_{RL} & = - \mmE_{P(\tau;\theta)} [R(\tau) \ \mathrm{log}P\left(\tau;\theta\right)], \\
    & \approx - \frac{1}{B} \sum_{b=1}^{B} R(\tau^{(b)}) \ \mathrm{log}P\left(\tau^{(b)};\theta\right),
    \end{aligned}
\end{equation}
\xgh{where $B$ is the training batch size. Now, we introduce these reward modules in detail.}

\subsubsection{Quality reward.}
Given a generated paragraph $\tau$, we can calculate some quantitative metrics for it, such as BLEU\cite{10.3115/1073083.1073135}, METEOR~\cite{Denkowski2014MeteorUL}, CIDEr~\cite{Vedantam2015CIDErCI}, \textit{etc}. Directly using these metrics as the training reward can boost the sentence generation quality.
In this paper, we simply adopt the BLEU score $R_1=\mathrm{BLEU}(\tau)$ as the reward since the BLEU score is one of the most popular automated and inexpensive metrics. In practice, the BLEU can be replaced with any metric that needs to be optimized.

\subsubsection{Adversarial reward.}
Based on a paragraph $\tau$, this module acts as a discriminator to determine whether $\tau$ is manual annotation (Real) or generated by the machine (Fake).
Following \cite{Yu2017SeqGANSG}, we use Convolutional Neural Network (CNN) to extract text features since it can capture sequence information and has shown \hmg{exhibited} high performance in the complicated sequence classification task \cite{Zhang2015TextUF}. 
Specifically, given a generated paragraph $\tau$, we first concatenate the token embedding as the text representation. 
\hmg{We then }use different numbers of kernels with different window sizes to extract different features over the text representation and produce a new feature map.
After applying a max-pooling operation, we perform a fully connected layer with Sigmoid activation to output a probability, which denotes the probability \hmg{that} the input text is real. The calculation can be formulated as $R_2 = \mathrm{CNN} (\tau)$.
Inspired by adversarial training \cite{cao18a}, this module aims to minimize the performance gap between humans and the \textit{Writer}.

\subsubsection{Alignment reward.}

A paragraph $\tau$ is supposed to align its enriched KG $\mG$ since $\tau$ is generated by \textit{Writer} according to the $\mG$. In this sense, we propose to compute the similarity between $\tau$ and $\mG$ based on the attention mechanism. Given an abstract $\tau$ with $T$ words, we first use Long Short-Term Memory (LSTM) to extract text representation $\bC=\{\bc_t\}$, where $ \bc_t\in \mmR^d$,  $t \in \{1,\dots,T\}$. Following AttnGAN~\cite{AttnGANFT}, we obtain the hidden representation as follows:
\begin{equation}
    \begin{aligned}
    \bq_t= \mathrm{Softmax} \left(\frac{(\bW_{Q} \bc_t) (\bW_{K} \bH)^\top}{\sqrt{d}}\right) \bW_{V} \bH,
    \end{aligned}
\end{equation}
where $\bW_{Q}$, $\bW_{K}$, \hmg{and} $\bW_{V}$ are trainable parameters, $\sqrt{d}$ is a scaling factor and $\bH \in \mmR^{d \times N}$ are node features obtained from the \textit{Writer}. With the help of \hmg{the} self-attention mechanism~\cite{Vaswani2017AttentionIA}, the hidden feature $\bq_t \in \mmR^{d}$ not only fuses the text representations but also merges graph information.
Then, we calculate the cosine similarity as matching score $R_3$ as follows:
\begin{equation}
    \begin{aligned}
    R_3 = \sum_{t=1}^{T} \frac{\bq_t^\top \ \bc_t}{\|  \bq_t\| \ \|  \bc_t\|}.
    \end{aligned}
\end{equation}
Thus far, we can obtain the rewards $R_{1}$,  $R_{2}$, \hmg{and}
$R_{3}$ from above the \textit{Reviewer} modules.
Finally, to train our DRAW network, we define the overall training loss as follows:
\begin{equation}
    \begin{aligned}
    \label{eq:overall_loss}
    \mathcal{L} = \mathcal{L}_{SL} + \lambda_{RL} \ \mathcal{L}_{RL},
    \end{aligned}
\end{equation}
where $\lambda_{RL}$ is a trade-off parameter. $\mathcal{L}_{SL}$  trains the DRAW network within supervised learning while $\mathcal{L}_{RL}$ allows the DRAW network to explore diverse generation via reinforcement learning and evaluate the generation from multiple orientations.

\section{Experiments}
\subsection{Datasets}
\subsubsection{AGENDA dataset.} 
AGENDA is one of the most popular KGs-to-text datasets, which concludes 40,000 pair samples collected from the proceedings of 12 top AI conferences.
Each sample consists of a title, an abstract, and the corresponding KG, which is extracted by the SciIE system.
The KG is composed of recognized scientific terms and their relationships.
In particular, the types of scientific terms include Task, Metric, Method, Material, and Other. \hmg{The} types of relationships include Used-for, Conjunction, Feature-of, Part-of, Compare, Evaluate-for, and Hyponym-of.

\subsubsection{M-AGENDA dataset.} 
To further demonstrate the effectiveness of our DRAW network, we create a new dataset, called M-AGENDA. 
Specifically, we first calculate the cosine similarity between each abstract and the others in the AGENDA dataset.
We select two most-related instances for each one and combine these three as a new data example in the M-AGENDA dataset.

\subsection{Experimental Settings}
\subsubsection{Implementation details.} 
Our DRAW network consists of three well-design modules, \ie{\textit{Reader}, \textit{Writer} and \textit{Reviewer}.}
\hmg{We first train our \textit{Reader}, \textit{Writer} and \textit{Reviewer} on AGENDA dataset. Then, we use the trained \textit{Reader} and \textit{Writer} model on the M-AGENDA to generate novel paragraphs.}
To speed up convergence early in training, we adopt different pretraining strategies for each module. For the \textit{Reader},
we first use TransE~\cite{NIPS2013_5071} to train entity and relation embeddings. 
\hmg{We then} aggregate information passed from a 2-hop neighborhood to update the embedding of each node.
Following~\cite{Nathani2019LearningAE}, we use Adam optimization with an initial learning rate of 0.1. 
For the \textit{Writer}, we pre-train for 30 epochs with early stopping. Following~\cite{ribeiro2020modeling}, we use Adam optimization with an initial learning rate of 0.5. To ensure the generation effect, we set the maximum generation length to 430.
For the \textit{Reviewer}, we pre-train the adversarial module with SGD optimization and initialize a learning rate of 0.001. When pre-training the graph encoder of the alignment module, we use the same model and parameters of \textit{writer}. \hmg{In addition}, we systematically adjust the values of $\lambda_{AR}$ and $\lambda_{MR}$ to conduct several ablation studies.
We find that the experimental results of different coefficient combinations 
fluctuate only around 0.1, \hmg{causing} little effect on the results.
\textit{Writer}-\textit{Reviewer} obtains the best results with { $\lambda_{AR} = \lambda_{MR} = 2$. We set the trade-off parameter $\lambda_{RL}=1$}.
We implement our method with PyTorch.

\begin{table}
	\centering
	
	\begin{tabular}{l|c c c }
		\hline
	Model & BLEU & METEOR   &CIDEr
		\\ \hline
		GraphWriter 
		
		& 14.44    & 18.80  &28.30 \\ 
		GraphWriter+RBS  &15.17  & 19.59  &- \\
		Graformer  & 17.33  &21.43  &- \\
		CGE-LW     & 18.01    & 22.34  &33.06  \\ \hline
        \textit{Writer}-\textit{Reviewer} (\textbf{Ours}) & \textbf{19.60}    & \textbf{24.03}  &\textbf{45.21} \\ \hline
	\end{tabular}
	\caption{Quantitative evaluations of generation systems on the AGENDA dataset (higher is better).}
	\label{tab:Automatic Evaluations}
\end{table}

\begin{table}[t]
	\centering

	\begin{tabular}{l|c c}
		\hline
		\multirow{2}[0]{*}{Paragraph} & \multicolumn{2}{c}{Turing Test Results}\\
		\cline{2-3}
		 & Human & Machine             \\ \hline
		Written by Human & 68\%    & 32\% \\ 
		Written by \textbf{DRAW} & 48\%   & 52\% \\ \hline
	\end{tabular}
	\caption{Quantitative results of Turing test.}
	\label{tab:Turing}
\end{table}

\begin{table}[t]
	\centering
	
	\begin{tabular}{l|ccc}
		\hline
		Model & BLEU & METEOR   &CIDEr         \\ \hline
		\textit{Writer}  & 18.01    & 22.34   &33.06 \\
		\textit{Writer}+Adversarial  & 19.37    & 23.87   &39.30 \\ 
		\textit{Writer}+Alignment & 19.33    & 24.00   &43.49 \\ 
		\textit{Writer}+Quality & 19.50    & 24.03    &44.40 \\ \hline
		\textit{Writer}-\textit{Reviewer} (\textbf{Ours)}     &\textbf{19.60}    & \textbf{24.03}   &\textbf{45.21} \\ \hline
	\end{tabular}
	\caption{Ablation study for modules used in the \textit{Reviewer} on the AGENDA dataset.}
	\label{tab:reviewer}
\end{table}

\begin{table}[t]
	\centering
  	
	\resizebox{\textwidth}{!}
	{
	\begin{tabular}{l|ccc}
		\hline
		Model       & Grammar   & Coherence   & Informativeness    \\\hline
		PaperRobot       & 5.11      &4.95    &5.01     \\ 
		CGE-LW       & 6.77     & 6.29   & 6.57   \\ \hline
		\textbf{DRAW (Ours) }             &\textbf{7.63}     &\textbf{6.83}  &\textbf{7.10}   \\ \hline
	\end{tabular}
    \caption{Automatic evaluations results (higher is better).}
	
	\label{tab:group_human}
	}
\end{table}

\begin{table*}[t]
    \begin{center}
	\centering
	\begin{tabular}{l p{15cm}}
	\toprule
	
        Initial KGs & \textbf{41} entities, \textbf{18} relations: (global scene-level contextual information, PART-OF, spatial context recurrent convnet model) ; (wikipedia, USED-FOR, multilingual ner systems) ; (local image de-scriptors, CONJUNCTION, spatial configurations)
        \dots
        \\ \midrule
        
        PaperRobot & In this paper we propose a novel approach for multilingual \textcolor{orange}{named entity} recognition tasks  . The proposed method is based on semantic similarity measure that can be used to improve \textcolor{orange}{word retrieval} performance by using \textcolor{orange}{wikipedia} type of words from \textcolor{orange}{text documents} and then build an efficient query \textcolor{orange}{language model} which allows users with similar information between entities as clusters across different domains : part-of-speech tags are generated through each user 's document representation ; our knowledge base system was evaluated over state-of-the-art approaches trained object \dots   [\textit{covering 6 entities.}]
        
         \\ 
    		CGE-LW  &  in this paper ,\dots we propose a \textcolor{orange}{spatial context recurrent convnet model} to incorporate \textcolor{orange}{global scene-level contextual information} into a spatial context recurrent convnet model for \textcolor{orange}{object retrieval} .\dots, and \textcolor{red}{the contextual information from candidate boxes is used for object retrieval}. a positional language model that captures \textcolor{red}{contextual information from candidate boxes for object retrieval}. the proposed system is evaluated on the \textcolor{orange}{tac-kbp 2010 data},and the experimental results show that the proposed system can significantly improve the entity linking performance\dots [\textit{covering 21 entities.}]

  \\ 
		DRAW &  in this paper , we propose a novel approach to  \textcolor{blue}{entity linking$ ^{1}$} based on \textcolor{blue}{statistical language model-based information retrieval$ ^{1}$} , which exploits both \textcolor{blue}{local contexts and global world knowledge$ ^{2}$} to improve the \textcolor{blue}{entity linking$ ^{2}$} performance.\dots, we propose a \textcolor{orange}{spatial context recurrent convnet model} to integrate \textcolor{orange}{global context features} with local \textcolor{blue}{image de-scriptors$ ^{3}$} ,\textcolor{orange}{ spatial configurations} , and \textcolor{blue}{global scene-level contextual information$ ^{3}$} into a spatial context recurrent convnet model\dots, and a recurrent network with \textcolor{orange}{local and global information} to guide the search for \textcolor{orange}{candidate boxes} for \textcolor{orange}{object retrieval}\dots [\textit{covering 26 entities.}]
  \\ \bottomrule
	\end{tabular}
	\caption{ Example outputs of various models. \xgh{To better visualize the generated text, we omit information irrelevant to the comparisons.} Repetitive words are represented in \textcolor{red}{red} and entities included in KGs are represented in \textcolor{orange}{orange}. The potential knowledge is represented in \textcolor{blue}{blue} with the corresponding superscript.
	}
	\label{tab:example}
	\end{center}
\end{table*}

\subsubsection{Evaluation metrics.} 
To demonstrate the quality of the generated paragraphs, we report both quantitative results and human study results.
We divide our evaluation into two parts: KGs-to-text evaluation and overall performance evaluation.

For KGs-to-text evaluations, we adopt three general quantitative evaluation metrics, \ie BLEU~\cite{10.3115/1073083.1073135}, METEOR~\cite{Denkowski2014MeteorUL} and CIDEr~\cite{Vedantam2015CIDErCI} to 
evaluate our \textit{Writer}-\textit{Reviewer}. In addition, to demonstrate the realness of the paragraphs generated by our model,
we also set up a Turing test. Specifically, we randomly select 100 abstracts and shuffle them to \hmg{find} an evaluation set, where half of the abstracts are written by authors and the rest are generated by our \textit{Writer}-\textit{Reviewer}. 
After that, we test the turkers on Amazon Mechanical Turk (AMT) to determine whether the paragraphs in the evaluation set are written by humans.

For overall performance evaluation, 
we set up a human study to rate the abstracts generated by DRAW network, CGE-LW and PaperRobot. 
For each model, we randomly select 50 generated paragraphs and score them in terms of {`grammar', `informativeness', and `coherence' on Amazon Mechanical Turk (AMT)}. 
Specifically, the metric `grammar' measures the paragraphs written in well-formed English. The metric `informativeness' denotes whether the paragraphs make use of appropriate scientific terms. The metric `coherence' denotes that the generated text conforms to general specifications. For example, a complete abstract should include a brief introduction \hmg{to} a task, describe the solution, analyze and discuss the results, and so on.
Each metric \hmg{described above}, contains 10 levels, \hmg{with} rankings from 1 to 10 (from bad to good). 

\hmg{Following the relation prediction task~\cite{Nathani2019LearningAE}, we evaluate our link prediction  method of \textit{Reader} on the proportion of correct entities in the top N ranks (Hits@N) for N=1,3, and 10.}

\subsection{KGs-to-text Evaluation on AGENDA Dataset}
To verify our model on KGs-to-text task, we compare our \textit{Writer}-\textit{Reviewer} against several state-of-the-art models including GraphWriter \cite{KoncelKedziorski2019TextGF}, GraphWriter+RBS \cite{An2019RepulsiveBS}, Graformer \cite{Schmitt2020ModelingGS} and CGE-LW \cite{ribeiro2020modeling} on the AGENDA dataset. 

\subsubsection{Results.} 
We report the results of our method and other compared models \hmg{with respect to} three quantitative evaluation metrics in Table~\ref{tab:Automatic Evaluations}. As shown in Table~\ref{tab:Automatic Evaluations},
our \textit{Writer}-\textit{Reviewer} \wzq{achieves better performance than all the compared models in three quantitative evaluation metrics.
Specifically, our \textit{Writer}-\textit{Reviewer}} outperforms the 
state-of-the-art method
CGE-LW by 1.6 points in BLEU, 1.7 points in METEOR and 12.2 points in CIDEr. 
These results demonstrate the superiority of our \textit{Writer}-\textit{Reviewer} in the KGs-to-text task.

In addition, we carry out a human evaluation to demonstrate the effectiveness of our \textit{Writer}-\textit{Reviewer}. 
To be specific, for each paragraph in the evaluation set, we ask the human to choose whether these paragraphs are written by human-authors.
From these results in Table~\ref{tab:Turing}, nearly half of the paragraphs generated by our \textit{Writer}-\textit{Reviewer} are reviewed as written by humans. More critically, 32\% of the paragraphs written by humans are chosen as written by the AI system. These results demonstrate that our \textit{Writer}-\textit{Reviewer} can generate \hmg{realistic paragraphs similar to those written by humans}.

\begin{table}[t]
	\centering

	\begin{tabular}{l|cccc}
	\hline
		\multirow{2}[0]{*}{Method} &
		 \multicolumn{3}{c}{Hits@N} \\
		 \cline{2-5}
	     & @1 & @3 & @10 &   \\
	 \hline
	    PaperRobot   &11.9  &19.5  &42.4    \\
		\textbf{Our}   &\textbf{36.8}  & \textbf{46.0}   & \textbf{56.1} 
		 \\
     \hline
	\end{tabular}
	
	\caption{Accuracy of the link prediction on the M-AGENDA dataset. Hits@N values are in percentage.}
	
	\label{tab:automatic_metric}
	
\end{table}
\subsubsection{Ablation studies in \textit{Reviewer}.}
To investigate the effect of different modules in 
\textit{Reviewer}, we conduct an ablation study.
As shown in Table~\ref{tab:reviewer},
\textit{Writer} combined with one of the modules in \textit{Reviewer} arbitrarily obtains better performance than \textit{Writer}, which demonstrates the effectiveness of the modules in \textit{Reviewer}.  \textit{Writer} combined with all the modules in \textit{Reviewer}, namely \textit{Writer}-\textit{Reviewer}, achieves best performance.

\subsection{Evaluation \wzq{on M-AGENDA Dataset}}

To show the effectiveness of our DRAW network, we conduct experiments on the M-AGENDA dataset. Since the M-AGENDA dataset does not provide ground-truth, we conduct human study instead of quantitative evaluations. Specifically, for each metric in the human study, we average the scores of the paragraphs rated by the humans as the final score.

\subsubsection{Results of DRAW.}
We report the experimental results of our DRAW network and other compared methods in Table~\ref{tab:group_human}. 
From these results, our DRAW network achieves the best performance in terms of `grammar', `coherence', and `informativeness'. 
Specifically, 
PaperRobot~\cite{Wang2019PaperRobotID} obtains poor performance due to the neglect of the topological structure between entities.
CGE-LW~\cite{ribeiro2020modeling} takes advantage of the graph information effectively and achieves 6.77, 6.29, and 6.57 points in terms of three metrics, but it also ignores the fact that the generated paragraphs are supposed to match the KGs.
Different from the methods above, our DRAW network not only performs link prediction with multi-hop information in the \textit{Reader} but also matches the graphs and the generated paragraphs, and thus achieves the best performance. 
More ablation experiments about \textit{Reader} can be found in the supplementary material.

\subsubsection{Results of \textit{Reader}.}
\hmg{
As shown in Table \ref{tab:automatic_metric}, we report the experimental results of the link prediction method of our \textit{Reader} and PaperRobot. Our method achieves the Hits@1, Hits@3, Hits@10 scores of 36.8, 46.0, and 56.1, outperforming the PaperRobot by 24.5, 26.5, and 13.9 points, respectively. 
It demonstrates the effectiveness of our link prediction method.
}

\subsubsection{Visualization analysis.}

As shown in Table~\ref{tab:example}, we visualize a generated paragraph of our DRAW network. 
\hmg{More visualization results can be found in the supplementary material.}
We see that our DRAW network has the ability to cover more entities (represented in \textcolor{orange}{orange}), while PaperRobot  mentions less entities in the given KG. In addition, CGE-LW tends to repeat unrelated entities/sentences (represented in \textcolor{red}{red}).
With the help of \textit{Reviewer}, the generated text of DRAW network is fluent and grammatically correct. Moreover, our DRAW network is able to discover the potential relationships between entities (represented in \textcolor{blue}{blue} superscript.)

\section{Conclusions and Future Work}
In this paper, we propose a Deep ReAder-Writer (DRAW) network that reads multiple AI-related abstracts and then writes a new paragraph to represent enriched knowledge combining the potential knowledge
covering the topics mentioned in the source abstracts. 
Inspired by the review process, we propose a
\textit{Reviewer} to rate the quality of the generated texts from different dimensions, which \hmg{serve} as feedback signals to refine our DRAW network.
Ablation experiments demonstrate the effectiveness of our method. Moreover, \textit{Writer}-\textit{Reviewer} achieves state-of-the-art results on KGs-to-text generation task.
In terms of human study, some generations of our DRAW network successfully pass the Turing test and confuse the turkers. In future study, we will extend the DRAW network to write a complete paper in an iterative manner and develop more techniques to discover novel ideas, such as creating new entities.

\section*{Acknowledgments}
This work was partially supported by 
Key-Area Research and Development Program of Guangdong Province 2018B010107001,
National Natural Science Foundation of China (NSFC) 61836003 (key project),
Program for Guangdong Introducing Innovative and Entrepreneurial Teams 2017ZT07X183,
International Cooperation open Project of State Key Laboratory of Subtropical Building Science, South China University of Technology (2019ZA01),
Fundamental Research Funds for the Central Universities D2191240.

\bibliography{DRAW}

\onecolumn
\appendix

\end{document}


\onecolumn
\appendix

\renewcommand{\thetable}{S\arabic{table}}
\renewcommand{\thefigure}{S\arabic{figure}}

\begin{center}
 {
  \Large{\textbf{Supplementary for ``How to Train Your Agent to Read and Write?''}}
 }
\end{center}

~\\

In the supplementary materials, we provide more experiment settings and results. 
We organize our supplementary materials as follows. 
In Section \ref{sec:supp_link_prediction}, we provide qualitative and quantitative results to demonstrate the effectiveness of the link prediction of our \textit{Reader}. 
In Section \ref{sec:supp_more_visualization}, we provide more visualization results to verify that our DRAW network outperforms considered baselines on the AGENDA and M-AGENDA datasets.

\section{Effectiveness of Link Prediction} \label{sec:supp_link_prediction}

\subsection{Visualization Analysis of Link Prediction}
To demonstrate the effectiveness of our method, we show the link prediction results of the \textit{Reader}, as shown in Figure \ref{fig:reader comparison}. 
We show the sub-graphs of the knowledge graphs enriched by \textit{Reader}.
For better visualization, we present new links with the dotted lines. Then, we see that our \textit{Reader} is able to obtain information and capture potential relationships.
For example, 'Spanish' is used for 'sentiment analysis'.

\begin{figure*}[h]
\centering{
    \includegraphics[width=1 \linewidth]{figures/reader_1.pdf}
    \includegraphics[width=1 \linewidth]{figures/reader_2.pdf}
    \includegraphics[width=1 \linewidth]{figures/reader_3.pdf}
    \caption{The sub-graphs of the knowledge graphs enriched by \textit{Reader}.
    The black solid lines are extracted by information extraction system SciIE. 
    The colored dotted lines represent the predicted relations.
    The reasonable relations are represented in dotted \textcolor{blue}{blue} lines. 
    }
    \label{fig:reader comparison}
}
\end{figure*}

\newpage
\subsection{Accuracy of Link Prediction}
Following the relation prediction task~\cite{Nathani2019LearningAE}, 
we evaluate our link prediction method on the proportion of correct entities in the top N ranks (Hits@N) for N=1, 3 and 10 compared with PaperRobot~\cite{Wang2019PaperRobotID}. 
For each datapoint on the M-AGENDA dataset, we randomly replace head and tail entities to construct negative samples and assign a score for each sample. Based on the score, we rank all samples to calculate the metric Hits@N.
As shown in Table \ref{tab:automatic_metric}, we demonstrate the effectiveness of our link prediction method, achieving the Hits@1, Hits@3, Hits@10 scores of 36.8, 46.0 and 56.1, outperforming the PaperRobot by 25, 26 and 14 points, respectively.

\begin{table}[h]
	\centering

	\begin{tabular}{l|cccc}
	\hline
		\multirow{2}[0]{*}{Method} &
		 \multicolumn{3}{c}{Hits@N} \\
		 \cline{2-5}
	     & @1 & @3 & @10 &   \\
	 \hline
	    PaperRobot   &11.9  &19.5  &42.4    \\
		\textbf{Our}   &\textbf{36.8}  & \textbf{46.0}   & \textbf{56.1} 
		 \\
     \hline
	\end{tabular}
	\caption{Accuracy of the link prediction on M-AGENDA. Hits@N values are in percentage.}
	\label{tab:automatic_metric}
	
\end{table}

\subsection{Ablation Study on Link Prediction}

We conduct an ablation study to demonstrate the effectiveness of link prediction in our \textit{Reader}. 
Specifically, for each metric in the human study, we average the score of the paragraphs rated by the humans as the final score.
We report the experimental results of our DRAW network and DRAW network without link prediction in Table~\ref{tab:group_human}. 
It shows that our DRAW network achieves the best performance in terms of `Grammar', `Coherence', and `Informativeness' and our DRAW network without link prediction has a lower score, which demonstrates the effectiveness of the modules in \textit{Reader}.

\begin{table}[h]
	\centering

	\begin{tabular}{l|ccc}
		\hline
		Model       & Grammar   & Coherence   & Informativeness    \\\hline
		DRAW (w/o link prediction)     & 6.58     & 5.50   & 5.86   \\ 
		\textbf{DRAW (Ours) }             &\textbf{7.63}     &\textbf{6.83}  &\textbf{7.10}   \\ \hline
	\end{tabular}
	\caption{Average ranking all paragraphs generated on the M-AGENDA dataset (higher is better).}
	\label{tab:group_human}
	
\end{table}

\section{More Visualization Results of DRAW} \label{sec:supp_more_visualization}

\subsection{More Comparison Results on M-AGENDA}

To demonstrate the effectiveness of our DRAW on the M-AGENDA dataset, we visualize more generated paragraphs of our DRAW network.
As shown in Table~\ref{tab:example}, our DRAW network is able to cover more entities (represented in \textcolor{orange}{orange}), while PaperRobot~\cite{Wang2019PaperRobotID} mentions less entity in the given knowledge graph. Besides, CGE-LW~\cite{ribeiro2020modeling} tends to repeat unrelated entities/sentences (represented in \textcolor{red}{red}).
With the help of \textit{Reviewer}, the generated text of our DRAW network is fluent and grammatically correct. Moreover, our DRAW network is able to discover the potential relationships between entities (represented in \textcolor{blue}{blue} with superscript).

\subsection{Results on Training More Paragraphs}

To further demonstrate the effectiveness of our DRAW network,  we first calculate the cosine similarity between each abstract and the others in the AGENDA dataset. 
After ranking by the similarity scores, we select five most-related instances for each one and combine then  as a new data instance. And then, we obtain a new dataset.
Then, we input the new dataset into our DRAW network to write a new paragraph that covers the topics mentioned in the input. As shown in Table~\ref{tab:example_5}, we visualize the generated paragraphs of our DRAW network. 
Similar to the experimental results on M-AGENDA dataset,  our DRAW is able to cover more entities (represented in \textcolor{orange}{orange}), discover the potential relationships (represented in \textcolor{blue}{blue} with corresponding superscript).

\begin{table*}[t]
    \begin{center}
	\centering
	\begin{tabular}{l p{15cm}}
	\toprule
	
        Initial KGs & \textbf{46} entities, \textbf{25} relations: (constrained minimization approach, USED-FOR, classification error) ; (constrained minimization approach, USED-FOR, classification accuracy) ; (classifier	, CONJUNCTION,constrained minimization approach)
        \dots
        \\ \midrule
        
        PaperRobot &  In this paper we propose a novel method for topic identification based on \textcolor{orange}{support vector machines}  . The proposed \textcolor{orange}{constrained minimization approach} is used to improve language classification performance with two different methods : (1) it can be viewed as an extension of \textcolor{orange}{discriminative training} data and its \textcolor{orange}{accuracy} by using gaussian mixture models that have been successfully applied against other state-of-the-art approaches from various domains including feature selection techniques are \dots   [\textit{Covering 6 entities.}]
        
         \\ 
    		CGE-LW  &  in this paper , we propose a novel approach to enhance the \textcolor{orange}{classifier accuracy} by incorporating the \textcolor{orange}{discriminative training} of \textcolor{red}{support vector machine and support vector machine} .\dots specifically , we use the generalized linear discriminate sequence kernel to build the vector support vector framework for the binary identification tasks . \dots the proposed approach is evaluated on the \textcolor{orange}{switchboard databases} and the results show that the proposed method outperforms the \textcolor{orange}{baseline classifiers} in terms of relative error reduction accuracy .\dots \textcolor{red}{compared with the state-of-art svm system} , the proposed pair-wise probability estimation algorithm is also shown to outperform \textcolor{red}{the state-of-art svm system} \dots [\textit{Covering 18 entities.}]

  \\ 
		DRAW & in this paper , \dots we propose a \textcolor{orange}{discriminative training approach} to improve the \textcolor{orange}{classification error classification accuracy} by incorporating the \textcolor{blue}{constrained minimization approach$ ^{1}$} with the \textcolor{blue}{latent semantic indexing matrix$ ^{1}$} and \textcolor{orange}{support vector machines} . \dots in order to deal with the large amount of training data , we use the \textcolor{orange}{generalized linear discriminate sequence kernel$ ^{2}$} to train the \textcolor{orange}{cepstral model$ ^{2}$} , which is then used to train a  logistic regression model . the proposed method is evaluated on the \textcolor{orange}{switchboard databases} and compared against \textcolor{orange}{the state-of-art svm system}\dots  moreover , it is also shown that the proposed method can reduce the number of training data required to achieve the same performance as the \textcolor{orange}{baseline classifiers} . finally , the proposed method is applied to the \textcolor{orange}{banking call routing system}\dots [\textit{Covering 22 entities.}]
		
  \\ \bottomrule
  \toprule
	
        Initial KGs & \textbf{38} entities, \textbf{15} relations: (orthogonal basis clustering, USED-FOR, latent cluster information) ; (pre-computed local structure information, USED-FOR, unsupervised feature selection methods) ; (unlabeled data	, USED-FOR,feature selection)
        \dots
        \\ \midrule
        
        PaperRobot &  In this paper we propose a novel approach for \textcolor{orange}{dynamic scene segmentation} based on the appearance model  . Our method is able to track moving objects from motion regions and can be used as an object tracking problem with latent variable flow . Depth of layered images captured by using particle filters that are estimated through each frame sequence which allows both \textcolor{orange}{global shape information} between consecutive frames per pixel or image boundaries into two views : their spatial temporal dynamics may fail at different locations than previous methods ; however they require accurate estimates over time varying lighting changes\dots [\textit{Covering 3 entities.}]
        
         \\ 
    		CGE-LW  &  in this paper , we address the problem of ranking a \textcolor{orange}{large-scale non-convex implementation} using \textcolor{orange}{multiple collaborative filtering datasets} . most existing \textcolor{orange}{unsupervised feature selection methods} rely on \textcolor{red}{pre-computed local structure information} in the form of \textcolor{red}{pre-computed local structure information} in the form of pairwise preferences .\dots  specifically , we formulate the \textcolor{red}{objective function} as a minimization factored form of the \textcolor{red}{objective function} subject to the gradient potentials of the \textcolor{red}{objective function} ,\dots , we incorporate the \textcolor{orange}{pairwise information} into the objective function , which can be solved efficiently by convex optimization . in addition , we propose a \textcolor{orange}{rank r score matrix} to capture the \textcolor{orange}{latent cluster information} from the \textcolor{orange}{unlabeled data} , which can be efficiently solved by the alternating minimization of the objective function\dots [\textit{Covering 14 entities.}]

  \\ 
		DRAW & in this paper , we consider the \textcolor{orange}{collaborative ranking setting} , where the goal is to learn a \textcolor{blue}{rank r score matrix$ ^{1}$} for \textcolor{blue}{pairwise data$ ^{1}$} .\dots however , in the collaborative ranking setting , the \textcolor{orange}{pairwise preferences} are not directly related to the \textcolor{orange}{pairwise information} , which is usually ignored in practice . \dots to solve the \textcolor{orange}{objective function} , we propose an efficient \textcolor{orange}{optimization algorithm} to solve the proposed objective function . moreover , we propose a \textcolor{blue}{regularized regression-based formulation$ ^{2}$} to optimize the \textcolor{blue}{objective function$^{2}$} by \textcolor{orange}{minimizing} the sum of the objective function subject to the common conditional potentials . extensive experiments on three \textcolor{orange}{real world datasets} demonstrate that our prunfs approach significantly outperforms the state-of-the-art unsupervised feature selection method in ranking . moreover , we also propose a \textcolor{orange}{large-scale non-convex implementation} based on the proposed orthogonal basis clustering \dots [\textit{Covering 16 entities.}]
  \\ \bottomrule
  
	\end{tabular}
	\caption{ Example outputs of various models on M-AGENDA with three paragraphs. To better visualize the generated text, we omit information irrelevant to the comparisons. Repetition words are represented in \textcolor{red}{red} and entities included in KGs are represented in \textcolor{orange}{orange}. The potential knowledge is represented in \textcolor{blue}{blue} with the corresponding superscript.
	}
	\label{tab:example}
	\end{center}
\end{table*}

\begin{table*}[t]
    \begin{center}
	\centering
	\begin{tabular}{l p{15cm}}

  \toprule
	
        Initial KGs & \textbf{21} entities, \textbf{8} relations: (stereo images, USED-FOR, 3d surface reconstruction) ; (direct method, USED-FOR, estimation of the vertex depths) ; (direct method, CONJUNCTION,3d surface reconstruction)
        \dots
        \\ \midrule
    
		DRAW & in this paper , we propose a novel method for piecewise-planar surface reconstruction from \textcolor{orange}{stereo images} . the proposed method is based on the minimization of an epipolar \textcolor{orange}{regularizer} on the 3d mesh , which is defined as the sum of a finite element model of a mirrored surface from curvature constraints on surfaces . the regularization is based on the \textcolor{blue}{finite element method$^{1}$} , which has been shown to be very effective for \textcolor{blue}{3d surface reconstruction$^{1}$} from stereo images . the key idea of our method is to use a \textcolor{blue}{finite element model$^{2}$} to represent the \textcolor{blue}{vertex estimation of the depths$^{2}$} in stereo images . in order to solve the problem , we propose a direct and efficient method to solve the problem of 3d shape from a single image of a mirrored surface from curvature constraints . the proposed direct method is based on the assumption that the epipolar field on surfaces is a \textcolor{orange}{triangular mesh} , and the \textcolor{orange}{epipolar constraint} on the 3d mesh is not needed . the proposed direct method has two main advantages : ( 1 ) it does not require the computation of a 3d mesh ; ( 2 ) it does not require the computation of a \textcolor{orange}{numerical solution} ; ( 3 ) it does not require the computation of a 3d \textcolor{orange}{mesh}\dots the proposed direct method is compared with the direct method using \textcolor{orange}{real and synthetic images} . the experimental results show that the proposed direct method is superior to the state-of-the-art methods with the direct method .[\textit{Covering 12 entities.}]
  \\ \bottomrule
  
	\end{tabular}
	\caption{ Example output of our model on M-AGENDA with five paragraphs. To better visualize the generated text, we omit information irrelevant to the comparisons.  Entities included in KGs are represented in \textcolor{orange}{orange}. The potential knowledge is represented in \textcolor{blue}{blue} with the corresponding superscript.
	}
	\label{tab:example_5}
	\end{center}
\end{table*}

\clearpage
\bibliography{RWR}